\documentclass[sigconf]{acmart}
\renewcommand\footnotetextcopyrightpermission[1]{} 
\usepackage{float}
\usepackage{times}  
\usepackage{helvet} 
\usepackage{courier}  
\usepackage{amsthm}
\usepackage{amsmath}
\usepackage{amssymb}

\def\BibTeX{{\rm B\kern-.05em{\sc i\kern-.025em b}\kern-.08em
    T\kern-.1667em\lower.7ex\hbox{E}\kern-.125emX}}

\AtBeginDocument{%
  \providecommand\BibTeX{{%
    \normalfont B\kern-0.5em{\scshape i\kern-0.25em b}\kern-0.8em\TeX}}}

\acmConference[KDD'20]{Proceedings of
the 26th ACM SIGKDD Conference on Knowledge Discovery and Data Mining}{August 23--27, 2020}{Virtual Event, CA, USA}
\acmBooktitle{Proceedings of
the 26th ACM SIGKDD Conference on Knowledge Discovery and Data Mining (KDD ’20), August 23--27, 2020, Virtual Event, CA, USA}

\begin{document}

\title{Reinforced Epidemic Control: Saving Both Lives and Economy}

\author{Sirui Song$^{\dagger\ddagger}$, Zefang Zong$^{\dagger}$, Yong Li$^{\dagger}$, Xue Liu$^{\ddagger}$, Yang Yu$^{\dagger}$}

\affiliation{%
  \institution{$^{\dagger}$ Tsinghua University $^{\ddagger}$ McGill University}
}



\email{yangyu1@tsinghua.edu.cn}

\begin{abstract}
   Saving lives or economy is a dilemma for epidemic control in most cities while smart-tracing technology raises people's privacy concerns. In this paper, we propose a solution for the life-or-economy dilemma that does not require private data. We bypass the private-data requirement by suppressing epidemic transmission through a dynamic control on inter-regional mobility that only relies on Origin-Designation (OD) data. We develop DUal-objective Reinforcement-Learning Epidemic Control Agent (DURLECA) to search mobility-control policies that can simultaneously minimize infection spread and maximally retain mobility. DURLECA hires a novel graph neural network, namely Flow-GNN, to estimate the virus-transmission risk induced by urban mobility. The estimated risk is used to support a reinforcement learning agent to generate mobility-control actions. The training of DURLECA is guided with a well-constructed reward function, which captures the natural trade-off relation between epidemic control and mobility retaining. Besides, we design two exploration strategies to improve the agent's searching efficiency and help it get rid of local optimums. 
   Extensive experimental results on a real-world OD dataset show that DURLECA is able to suppress infections at an extremely low level while retaining 76\% of the mobility in the city. Our implementation is available at \url{https://github.com/anyleopeace/DURLECA}.
\end{abstract}

%
%




\maketitle
\section{Introduction}
\begin{figure}[t]
    \centering
    \includegraphics[width=0.4\textwidth]{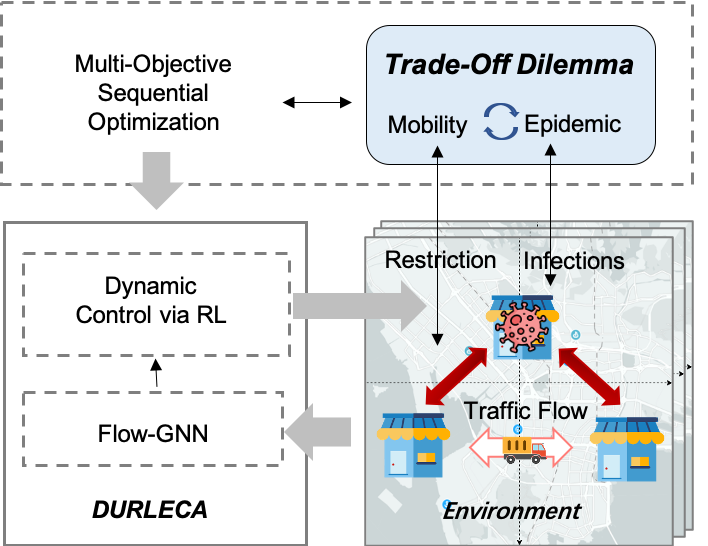}
    \caption{The overview of reinforced epidemic control system.}
    \label{fig:Overview}
    \vspace{-0.5cm}
\end{figure}

The epidemic has always been a threat to human society by exposing us in front of a dilemma between saving lives or economy. The virus infects gathering people and spreads through daily commute \cite{poletto2012heterogeneous, balcan2010modeling, wesolowski2012quantifying}. Controlling the spread of the virus must cut off daily mobility, which is a pillar of the modern economy. For instance, the recent outbreak of COVID-19 has caused millions of infections and hundreds of thousands of death tolls. The epidemic forces many municipal governments to issue  a stay-at-home order, which is a Fully LockDown (FLD) policy. FLD in most cities lasts for weeks thus deeply hurts the economy~\cite{barua2020understanding}. Some municipalities try to only quarantine symptomatic people and their close contacts at the early stage of the epidemic. However, this infected-individual-quarantine policy would be only implementable when governments are able to accurately and comprehensively trace risky people. It is also unreliable when there exist many asymptomatic infected people. Current computer-science explorations pursue using smartphone data to infer and trace highly-risky people \cite{ferretti2020quantifying, oliver2015mobile}. However, fully tracing individual mobility and contacts requires full coverage of smartphones and further raises the concern of threatening privacy~\cite{privacy}. According to an investigation by the University of Maryland and The Washington Post, around 60\% of respondents either prefer not sharing their private information or do not own a smart phone~\cite{timberg2020most}. 
In summary, the vast amount of complex individual mobility and asymptomatic infected people prevent current epidemic-control policies from cutting off virus spread without hurting the economy when private information cannot be fully captured.

We in this research demonstrate that a smart epidemic control policy is still available even if private mobility information is unavailable. We develop a dynamic control framework to avoid an epidemic outbreak by limiting the probability of risky mobility's occurrence. Instead of targeting and limiting risky individual's mobility according to private data, our framework estimates each urban region's risk of having a high infected population and uses the estimation to control inter-regional mobility. Highly-risky inter-regional mobility will be limited to suppress the probability of infected people's movement. Because the infected people are a small proportion of the population even in a seriously infected city, only a small number of mobility must be restricted. It is possible to avoid an epidemic outbreak by heterogeneously limiting little inter-regional mobility. Furthermore, the estimation is based on the regional aggregate demand for mobility and the regional epidemic statistics. Thus, private data is dispensable. 

However, there exist three specific challenges causing the complexity of estimating and controlling inter-regional mobility for suppressing the infection and protecting the economy. 
First, urban mobility is vast and temporally varying, making it hard to target the really risky mobility. 
Further, the requirements of the policy's practicality sophisticate the design of the epidemic-control policy. An implementable control policy cannot continuously quarantine the same urban region for too long. 
Last but not the least, the search for policy is difficult. Due to the exponentially-increasing nature of epidemics, the number of future infections is a highly non-convex function of each previous decision, making it hard to explore the policy space. Furthermore, the dual objectives cause the policy exploration often end up stuck in local optimums, which is also exacerbated by the non-convexity of infections.


With the consideration of the above challenges, we develop a DUal-objective Reinforcement-Learning Epidemic Control Agent (DURLECA) framework by combining Graph Neural Network (GNN) and Reinforcement Learning (RL) approach, to search out an effective mobility-control policy. 
DURLECA hires a GNN to estimate the virus-transmission risk induced by urban mobility, which is a dynamic flow on a graph. Based on the estimated risk, the RL agent periodically determines the extent of the restriction on each inter-regional mobility. 
The GNN of DURLECA is developed with a novel architecture, namely Flow-GNN, to fit the virus spread process on mobility flows, which existing GNN architectures are incompatible to characterize. 
We also carefully construct a reward function for the RL agent to precisely capture the natural trade-off relation between epidemic control and urban-mobility retaining. The reward function also considers the difference between continuous and intermittent restrictions on the same region. 
Furthermore, we develop two RL exploration strategies that appropriately incorporate epidemic expert knowledge for guiding and stabling policy exploration.

Supported by a Susceptible-Infected-Hospitalized-Recovered (SIHR) epidemic simulation environment developed from the traditional SIR model~\cite{SIR}, DURLECA is able to successfully search out a mobility-control policy that suppresses the epidemic and retains most of the mobility. Our experiments on a real-world mobility dataset collected in Beijing demonstrate the effectiveness of DURLECA. Even if the city starts to suppress an epidemic\footnote{The $R0$ is $0.19 \sim 1.08$ for SARS, $1.4 \sim 2.8$ for Influenza, $1.94 \sim 5.7$ for COVID-19, according to \url{https://en.wikipedia.org/wiki/Basic_reproduction_number}.} whose $R0=2.1$ after 20 days of discovering the first patient, DURLECA still finds out a policy where:
\begin{itemize}
    \item 
    The peak demand for hospitalization is under 1.3\textperthousand\footnote{The hospital bed density is 2.9 \textperthousand \text{ in} U.S., 4.2\textperthousand  \text{ in} China, and 13.4\textperthousand  \text{ in} Japan, according to \url{https://www.indexmundi.com/g/r.aspx?v=2227&l=en}.} of the whole population. The average demand for hospitalization is controlled under 0.4\textperthousand.  
    \item 76\% of the total mobility is retained. In more than $70\%$ intervened days, two-thirds regions retain over $70\%$ mobility. No region ever experiences a stringent day, i.e., daily retained mobility lower than 20\%. 
\end{itemize}

In summary, the contribution of this paper is in three-folds:
\begin{itemize}
   \item 
   We bypass the privacy concern for smart epidemic control. Instead of directly tracing and quarantining risky individuals, we suppress the risk of an epidemic outbreak by estimating and restricting risky inter-regional aggregate mobility. 
   \item
   We develop DURLECA to dynamically generate customized control actions for inter-regional mobility, which allows a smart solution for the life-or-economy dilemma of epidemic control.
  \item 
    We propose innovative approaches to guarantee DURLECA's capability. We design a novel GNN architecture that can fit the epidemic transmission dynamics. Our RL reward function captures the nature of the trade-off relation between epidemic suppressing and mobility retaining, and reflects practical requirements. We also develop two RL exploration strategies that appropriately incorporate epidemic expert knowledge for guiding and stabling policy exploration. 
\end{itemize}
\section{Preliminary and Problem Formulation}
During the stay-home order of COVID-19, governments distribute mobility quotas per day to each household for retaining the basic economic activities, such as people's procurement for food. According to the current quota regulation, we develop a new policy environment. We assume that the government periodically predicts or collects aggregate demands for inter-regional mobility of every Origin-Destination (OD) pair. The government also collects information about the number and location of current discovered patients. 
Those pieces of information are used to determine the quotas for each inter-regional mobility. 
The quota-distribution aims at minimizing the risk of epidemic break out in the foreseeable future periods while maximizing the mobility demands. We in this section present the modeling of mobility and epidemic that supports quota allocation.

\subsection{Mobility Modeling} 
We model a city's urban-mobility demand at time step $\tau$ as a mobility matrix $M_d^\tau$, whose element $M_{i,j}^\tau$ represents the inter-regional mobility demand, i.e., the number of people who demand to move, from $i$ to $j$. According to $M_d^\tau$ and the epidemic information, the city government determines a mobility quota matrix $p^\tau$ at $\tau$, whose element $p_{i,j}^\tau$ is the quota rate distributing to the mobility demand from $i$ to $j$. Therefore, the allowed inter-regional mobility denoted by $M_{p,i,j}^\tau$ is calculated according to the following equations.
\begin{align}
    & M_{p,i,j}^\tau = p_{i,j}^\tau M_{d,i,j}^\tau ,  \\
    & M_p^\tau = \mathcal{T}(M_d^\tau, p^\tau) = M_{d}^\tau \odot p^\tau,
\end{align}
where $\mathcal{T}$ refers to the mobility control function and  $\odot$ denotes for element-wise multiplication. Note that $M_d^\tau$, $M_p^\tau$, and $p^\tau$ are $K \times K$ matrices, where $K$ is the number of regions in the studied city. We summarize the mobility-related notations in Appendix.

\subsection{Epidemic Modeling} 
The main challenge for urban epidemic control comes from infected people who are infectious but asymptomatic. Therefore, we develop a new epidemic model to capture the difference between asymptomatic people and symptomatic people. Our model is based on the traditional Susceptible-Infected-Recovered ($SIR$) model in public-health literature \cite{SIR}. We introduce a new state beyond $SIR$ and denote it by Hospitalized ($H$). People in state $H$ are infected with symptoms and thus will be quarantined or hospitalized. They will not participate in urban mobility and will not contribute to new infections. 
We refer our model as $SIHR$ model\footnote{This modeling is different from the SEIR model \cite{SEIR} which assumes the asymptomatic people are not infectious.}. 

We use our $SIHR$ model to capture the dynamic process of infection spread over urban mobility. We denote region $i$'s epidemic state by $E^\tau_i = \{S^\tau_i,I^\tau_i,H^\tau_i,R^\tau_i\}$, whose each element respectively denotes the susceptible, infected, hospitalized, and recovered population of $i$ at $\tau$. We use $E_{v,i}^\tau =\{S^\tau_i+I^\tau_i,H^\tau_i,R^\tau_i\}$ to represent the visible state of $i$ at $\tau$, where the healthy people cannot be differentiated from infectious asymptomatic people. We denote the total population of $i$ at $\tau$ by $N_i^\tau$.

The epidemic state $E_i^\tau$ is updated in each time step. For each time step $\tau$, we separate $\tau$ into two sub-steps: mobility happens and infection occurs. At the mobility-happening sub-step, people accomplish their moves between regions. We use $E_i^{s,\tau}$ to represent the epidemic state of the staying people while $E_i^{m,\tau}$ represents the new arrival's. The overall epidemic state at the mobility-happening sub-step, denoted as $\hat{E}_i^{\tau}$, is calculated as follows:
\begin{align}
    & E_i^{s, \tau} = E_i^{\tau} - \sum_j \frac{M_{p,i,j}^{\tau}}{N_i^{\tau}}E_i^{\tau}, \label{eq:flow_stay}\\
    & E_i^{m, \tau} =  \sum_j \frac{M_{p,j,i}^{\tau}}{N_j^{\tau}}E_j^{\tau}, \label{eq:flow_movein}\\
    & \hat{E}_i^{\tau} = E_i^{s, \tau} + E_i^{m, \tau}. \label{eq:flow_end}
\end{align}

At the infection-occurring sub-step, people that stay at $i$ infect each other. Simultaneously, new arrivals at $i$ infect each other. Therefore, the epidemic state is updated as follows:
\begin{align}
    & S_i^{\tau+1} = \hat{S}_i^{\tau} - 
    \frac{\beta_i^{s,\tau} S_i^{s,\tau} I_i^{s,\tau}}{N_i^{s,\tau}} - 
    \frac{\beta_i^{m,\tau} S_i^{m,\tau} I_i^{m,\tau}}{N_i^{m,\tau}} , \\
    & I_i^{\tau+1} = \hat{I}_i^{\tau} +
    \frac{\beta_i^{s,\tau} S_i^{s,\tau} I_i^{s,\tau}}{N_i^{s,\tau}} +
    \frac{\beta_i^{m,\tau} S_i^{m,\tau} I_i^{m,\tau}}{N_i^{m,\tau}} - \gamma \hat{I}_i^{\tau}, \\
    & H_i^{\tau+1} = H_i^{\tau} + \gamma_i^\tau \hat{I_i}^{\tau} \label{eq:hospital} - \theta_i^\tau H_i^{\tau},\\
    & R_i^{\tau+1} = \hat{R_i^{\tau}} + \theta_i^\tau H_i^{\tau}, \label{eq:R}
\end{align}
where $\{\hat{S}_i^{\tau},\hat{I}_i^{\tau},\hat{H}_i^{\tau},\hat{R}_i^{\tau}\}$ are elements of $\hat{E}_i^{\tau}$. $\{ \beta^{s,\tau}_i, \beta^{m,\tau}_i \}$ are the epidemic's transmission rate for the staying people and the moving people respectively. $\gamma^\tau_i$ is the hospitalized rate and $\theta^\tau_i$ is the recover rate. We use one set of $\{\beta^{s}, \beta^{m}, \gamma,\theta \}$ for all regions at all time steps for simplification. We introduce how we estimate $R0$ in Appendix.

\subsection{Multi-Objective Sequential Control Problem Formulation}
The above mobility and epidemic modeling allow us to formulate the dynamic inter-regional mobility control problem for minimizing infections and maximizing mobility retaining, shown in Equation (\ref{eq:trans_model})$\sim$(\ref{eq:reward}):
\begin{align}
    & M_p^\tau = \mathcal{T}(M_d^\tau, p^\tau) \label{eq:trans_model}, \quad E_p^{\tau+1} = \mathcal{E}(M_p^\tau, {E}_p^\tau) \\
    & P^{t,T} = \mathrm{arg}\,\underset{P}{\mathrm{max}} \sum_{\tau=t}^T \mathcal{O}(M_P^\tau, E_P^\tau), \label{eq:reward}
\end{align}
where $\mathcal{O}$ is the objective function, satisfying $\partial^2 \mathcal{O}/\partial M_p \partial E_p < 0$ because of the trade-off nature between epidemic control and mobility retaining. $\mathcal{O}$ should also meet some practical requirements.
In the next section, we detail the design of the objective function and use it as the reward function of the RL module of DURLECA. Besides, we particularly consider the fact that the frequency of government interventions is lower than the frequency of mobility. Therefore, the mobility and infection updates per hour while the government determines mobility quotas per four hours.

\section{DURLECA}
\begin{figure}[t]
\vspace{-0.2cm}
    \centering
    \includegraphics[width=0.5\textwidth]{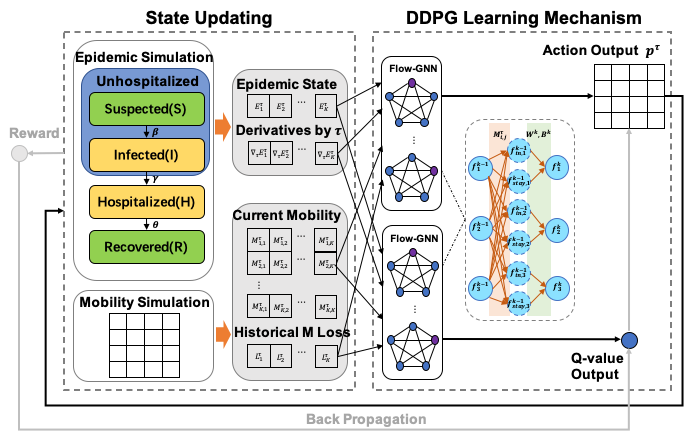}
    \caption{The details of the proposed DURLECA.} 
    \label{fig:system}
    \vspace{-0.4cm}
\end{figure}

DUal-objective Reinforcement-Learning Epidemic Control Agent (DURLECA) is a GNN-enhanced RL agent to estimate regional infection risk and determine mobility quota. An overview of DURLE-CA is shown in Figure~\ref{fig:system}.  At each time step $\tau$, DURLECA acquires an observation $E_v^\tau$ from the environment. According to $E_v^\tau$ and the demand mobility $\{M_d^\tau,M_d^{\tau+1},M_d^{\tau+2},M_d^{\tau+3}\}$, our RL agent gives a control action $p^\tau$ for the optimization problem in Equation (\ref{eq:reward}). 
In the rest of this section, we provide the details of DURLECA.

\subsection{Reinforcement Learning}

We now re-formulate the multi-objective sequential control problem using the basic factors of RL, i.e., state, action, reward, and learning algorithm.

\textbf{State:} We take the visible epidemic state $E_v^\tau$, its temporal one-order derivatives $\nabla_\tau E_v^\tau$, the mobility demand $\{M_d^\tau,M_d^{\tau+1},M_d^{\tau+2},M_d^{\tau+3}\}$, and the historical mobility loss $L^\tau$ (defined later) as the state for RL.
    
\textbf{Action:} The action of RL is defined as the mobility restriction $p^\tau$ determining the quota rate for each inter-regional mobility at $\tau$. Each element of $p^\tau$ is a real number between 0 and 1.

\textbf{Reward:} The reward function is designed to reflect the objective $\mathcal{O}$ of the optimization problem in Equation (\ref{eq:reward}). It includes two terms: an infection-spread-cost term and a mobility-restriction-cost term. In order to guide the RL agent to effectively find an effective and practical mobility-control policy. We design the reward function to satisfy the following three requirements:
\begin{itemize}
    \item Reflecting the trade-off relation between infection control and mobility retaining. 
    \item Capturing the exponential growth of the social cost caused by infection spread. The social cost is low when the infected population is small. However, the social cost will skyrocket once the infection population exceeds the capacity of the city's healthcare system. 
    \item Penalizing continuous mobility restrictions in the same region. People's tolerance for mobility restrictions is limited. Thus, the reward function has to include a growing penalty for continuously restricting the same region. 
\end{itemize}
According to the above three requirements, we separately design the infection-spread-cost term and mobility-restriction-cost term. We denote the infection-spread-cost by $R_h^\tau$ and model it as follows:
    \begin{align}
        R_h^\tau = k_h \text{exp}(\frac{\frac{1}{K}\sum_i H_i^\tau}{H_0}), \label{eq:infection}
    \end{align}
where $k_h$ is a hyper-parameter determining the start-up social cost of a city having the first patient while $H_0$ is a hyper-parameter determining how the social cost increase along with the number of patients. The mobility-restriction-cost term is denoted by $R_m^\tau$ and defined below:
\begin{align}
        & L_i^\tau = \sum_{t=0}^{\tau-1} \lambda^{\tau - t} \frac{M_{d,i}^{t}- M_{p,i}^{t}}{ \overline{M_{d,i}}}, \\
        & R_m^\tau = \frac{1}{K} \sum_i \text{exp}(\frac{L_i^\tau}{L_0}) \frac{M_{d,i}^{\tau}- M_{p,i}^{\tau}}{\overline{M_{d,i}}}. \label{eq:restriction}
    \end{align}
Here, $L_i^\tau$, the historical mobility loss, is the amount of mobility restricted in history and induces an exponentially-growing penalty $R_m^\tau$ on the current restriction. The hyper-parameter $\lambda$ determines the discount rate of historical restrictions' impacts. The hyper-parameter $L_0$ determines how large the penalty is for continuously limiting the same region. Finally, we develop the RL reward function $\mathcal{R}$ as follows:
    \begin{align}
        \mathcal{R}(M_p^{\tau}, E_p^{\tau}) = -(R_m^\tau + R_h^\tau).
    \end{align}
Note that our design enables the reward function to reflect that the infection-spread-cost booms once the whole city's hospitalized population exceeds the city's healthcare system capacity while the mobility-restriction-cost skyrockets if any single region is continuously restricted for multiple periods.

\textbf{Learning Algorithm:}
DURLECA employs a Deep Deterministic Policy Gradient (DDPG) \cite{DDPG} agent to search for mobility-control policy because the action space is continuous. The DDPG agent is composed of a critic network and an actor network. The critic network aims to estimate the expected reward gained by a control action. The actor network searches for the best action, which gives quota rates for all inter-regional mobility, by maximizing the critic network's output. We use Parameter Noise \cite{param_noise} to improve exploration during RL training.

\subsection{Flow-GNN}
Both critic and actor networks have to well capture the graph nature of urban mobility, where regions are nodes connected by OD flows. 
Therefore, we adopt GNN to develop both of them. We design a novel GNN architecture so that GNN can characterize the epidemic transmission process driven by regional infection aggregation upon inter-regional mobility. 
We refer our proposed GNN as Flow-GNN, which is developed on the basis of GraphSage~\cite{graphsage}.

In particular, we design Flow-GNN fit for the low-frequency mobility control associated with high-frequency mobility dynamics. Considering that we determine mobility quota per four hours, we include 4 Flow-GNN layers in our network and input edge information chronologically. The edge-input information for the $k$-th layer is $M_*^{\tau+k-1}$. We use $f_i^{k}$ to denote the feature of region $i$ outputted by the $k$-th GNN layer and calculate it according to the following equations:
\begin{align}
    & f_{stay,i}^{k-1} = (1-\sum_j \frac{M_{*,i,j}^{\tau+k-1}}{N_i^{\tau+k-1}})f_i^{k-1}, \label{eq:stay_feature}\\
    & f_{in,i}^{k-1} = \sum_j \frac{M_{*,j,i}^{\tau+k-1}}{N_j^{\tau+k-1}}f_j^{k-1}, \label{eq:in_feature} \\
    & f_i^k = \sigma (W^k (f_{in,i}^{k-1}, f_{stay,i}^{k-1}) + B^k). \label{eq:all_feature}
\end{align}
Here $(f_{in,i}^{k-1}, f_{stay,i}^{k-1})$ denotes for concatenation, $\sigma$ is a non-linear activation function, and $W^k, B^k$ are trainable parameters. Specifically, we input the first layer with $f_i^0 =\{E_{v,i}^\tau, \nabla_\tau E_{v,i}^\tau \}$. 

The above equations correspond to our modeling of epidemic transmission in Section 2.2, where we separate each time step $\tau$ into mobility-happened sub-step and infection-occurred sub-step. Equation (\ref{eq:stay_feature}) describes the epidemic feature of staying population at mobility-happened sub-step while Equation (\ref{eq:in_feature}) represents the new-arrival population's epidemic feature in the same sub-step. Equation (\ref{eq:all_feature}) characterizes the epidemic transmission in the staying population and the new-arrival population. 

\subsection{Exploration Strategies}
The exponentially-increasing nature of epidemics and our dual objectives cause difficulties for RL exploration and increase the agent's risk of falling into local optimums. 
We design two RL exploration strategies to address this problem. The first strategy is to incorporate pseudo-expert knowledge to improve RL searching efficiency. The second is to protect the agent from falling into local optimums by stopping it from exploring apparently unreasonable policies. 

\textbf{Generating-and-Incorporating Pseudo Expert:} We can generate simple but dynamic policies according to current epidemic-management experience, which can be a good start point for RL exploration. For instance, most cities currently restrict the mobility of regions with a large symptomatic population while a region has urgent reopening demand if it has been continuously locked down for a long time. Thus, we design a pseudo expert, which control $p_{i,j}^\tau$ as follows:
\begin{align}
    p_{i,j}^\tau =
    \begin{cases}
        0 & \quad \text{if } H_i^\tau > X_h \text{ and } L_i^\tau < X_l  \\
        1 & \quad \text{else} .
    \end{cases} \label{eq:expert}
\end{align}
The expert will lock a region down based on two conditions: 1) the number of hospitalized, or symptomatic patients in this region, exceeds the threshold $X_h$; 2) this region has not been restricted very much in history, reflected by that $L_i^\tau$ does not exceed the threshold $X_l$. During testing, this expert is also used as a comparing baseline. 

We let the agent first explore with expert's guidance and then gradually learn to explore by itself to outperform the expert. The idea is inspired by the approach adopted to develop AlphaGO~\cite{go}. Specifically, we set an adaptive probability for the agent to directly choose the expert action instead of taking an action by itself during training. This design enables the agent to compare the pseudo-export strategy with its own, which avoids the agent to move towards inefficient directions at the initial stage of training. The adaptive probability decreases along with training steps, which enables the agent to broadly explore and outperform the expert at the later stage of training.

\textbf{Avoiding Extreme Points:} The wide exploration might lead the agent to fall into some extreme points. The training might be unstable due to a sudden large loss caused by a poor control action. 
Meanwhile, the strong incentive of avoiding the large loss will force the agent to fall into local optimal control policies, such as a forever fully-lockdown. 
To avoid such extreme points, we set two rules:
\begin{itemize}
    \item The infection threshold $I_{t}$: If the agent explores into a state where the regional mean number of infected people exceeds $I_{t}$, it will end the episode and receive a large penalty.
    \item The lockdown threshold  $L_{t}$: If the agent explores into a state where there exists a region $i$ that $L_i^\tau$ exceeds $L_{t}$, it will end the episode and receive a large penalty.
\end{itemize}
The two rules are straightforward but effective to help the RL agent avoid potential local optimums.


\section{Experimental Evaluations}
In this section, we conduct extensive experiments to answer the following research questions:

\textbf{RQ1:} Can DURLECA resolve the life-or-economy dilemma? 

\textbf{RQ2:} Can DURLECA adapt to both early intervention and late intervention?

\textbf{RQ3:} Can DURLECA be generalized to different cities and different diseases?

Besides, we conduct ablation studies in Appendix to evaluate the effectiveness of our proposed Flow-GNN and RL exploration strategies.

\subsection{Dataset}
We use a real-world OD dataset collected by a mobile operator in Beijing to evaluate DURLECA. The dataset divides Beijing into $17 \times 19$ regions and covers 544,623 residents. Averagely, each region has $1686$ observed residents. The dataset covers 24-hour OD-flows for the whole month of January 2019. We repeat the one-month data 24 times and get a prolonged dataset of 24 months so that we have a sufficiently long period for discussing epidemic control. We list other details in Appendix.

\subsection{Metrics and Settings}
We design six metrics to evaluate the performance of DURLECA on resolving the life-or-economy dilemma. We introduce the metrics in the following and summarize them in Table \ref{tab:metrics}. 

\begin{table}[ht]
\vspace{-0.3cm}
\centering
\begin{tabular}{l|l|l}
\hline
\hline 
Metric & Value & Physical Meaning \\
\hline
Mean/Max $H$ & 0-1686 & Temporal mean/max of $H$\\
Total $R$ & 0-1686 & Total $R$ after the epidemic \\ \hline
$Q$ & 0-1 &  Total quota rate \\
$T_{20\%}^c$ & 0-744 & The city 20\%-mobility duration\\
$T_{20\%}^r$  & 0-744 & The region 20\%-mobility duration \\
\hline
\hline
\end{tabular}
\caption{The summary of metrics and related value ranges.}
\label{tab:metrics}
\vspace{-0.7cm}
\end{table}

We select three metrics to assess the epidemic-suppressing performance of an epidemic control policy, including the total number of infected people that is equal to $R$ at the end of the epidemic period, the mean number of hospitalized people whose value is the mean of $H$ over time, and the peak demand for hospitalization capacity that is equal to the max value of $H$ over time. Total $R$ determines the total social medical costs while both Mean/Max of $H$ reflect the sustained and peak pressure on the healthcare system. We also select three metrics to assess the mobility-retaining performance of an epidemic control policy, including the total ratio of retained mobility $Q$, the duration of stringent mobility restrictions on the whole city $T_{20\%}^c$, and the duration of stringent mobility restrictions on the most restricted region $T_{20\%}^r$.

\textbf{Epidemic Settings:} Without the loss of generality, we set $\beta_s = \frac{0.3}{24}, \beta_m=\frac{3}{24}, \gamma=\frac{0.3}{24}, \theta = \frac{0.3}{24}$ in most of our experiments. The estimated basic reproduction number $R_0$ is 2.1.

\textbf{Intervention time:} We define $t_{start}$ as the time when the policymakers discover the epidemic and start to intervene. In our experiments, we compare results with $t_{start}=0,10,20$.

For more details about our experiment settings, please refer to Appendix. 
Our implementation is available online at \url{https://github.com/anyleopeace/DURLECA}.

\subsection{Performance Comparison}

\textbf{Baselines:}
We set four different expert baselines to simulate different real-world expert policies and compare them with DURLECA on resolving the life-or-economy dilemma.
\begin{itemize}
    \item \textbf{EP-Fixed:} In the real world, a simple but inflexible control is to restrict all mobility in the city. For simulation, we design EP-Fixed to give a fixed quota rate $X_q$ to all inter-regional mobility during the whole epidemic period. We set  $X_q=\{0.15,0.2\}$ in our experiments, as we find them at the boundary of successfully controlling the epidemic.
    
    \item \textbf{EP-Soft:} We design an expert baseline following Equation (\ref{eq:expert}), which softly depends on the historical mobility loss $L_i^\tau$ and the current hospitalized population $H_i^\tau$ to determine whether to lock down a region. We set $X_h=0,X_l=168$ in our experiments. $X_h=0$ guarantees the expert receive equivalent information compared with DURLECA. $X_l=168$ corresponds to the real-world control policy in some countries: a continuous 7-day (168-hour) lockdown.
    
    \item \textbf{EP-Hard:} Without softly depending on the historical mobility loss, an expert can reopen a region if it has been locked down for successive $X_t$ days. This expert, namely EP-Hard, gives daily quota as follows:
    \begin{align}
    p_{i,j}^\tau &=
    \begin{cases}
        0 & \quad \text{if } H_i^\tau > X_h \text{ and } \sum_{t=1}^{X_t} M_{p,i}^{\tau-t} > 0  \\
        1 & \quad \text{else} .
    \end{cases} \label{eq:expert_hard}
    \end{align}
    We set $X_h=0,X_t=7$ for a similar reason of EP-Soft.
    
    \item \textbf{EP-Lockdown:} The most robust and conservative policy is to lock down the whole city until the epidemic ends. To simulate it, we design an expert following Equation (\ref{eq:expert}) but with $X_h=0,X_l=inf$. It can lock down a region for an any-long time until the hospitalized population becomes zero. 
\end{itemize}

\begin{figure}[t]
    \centering
    \includegraphics[width=0.46\textwidth]{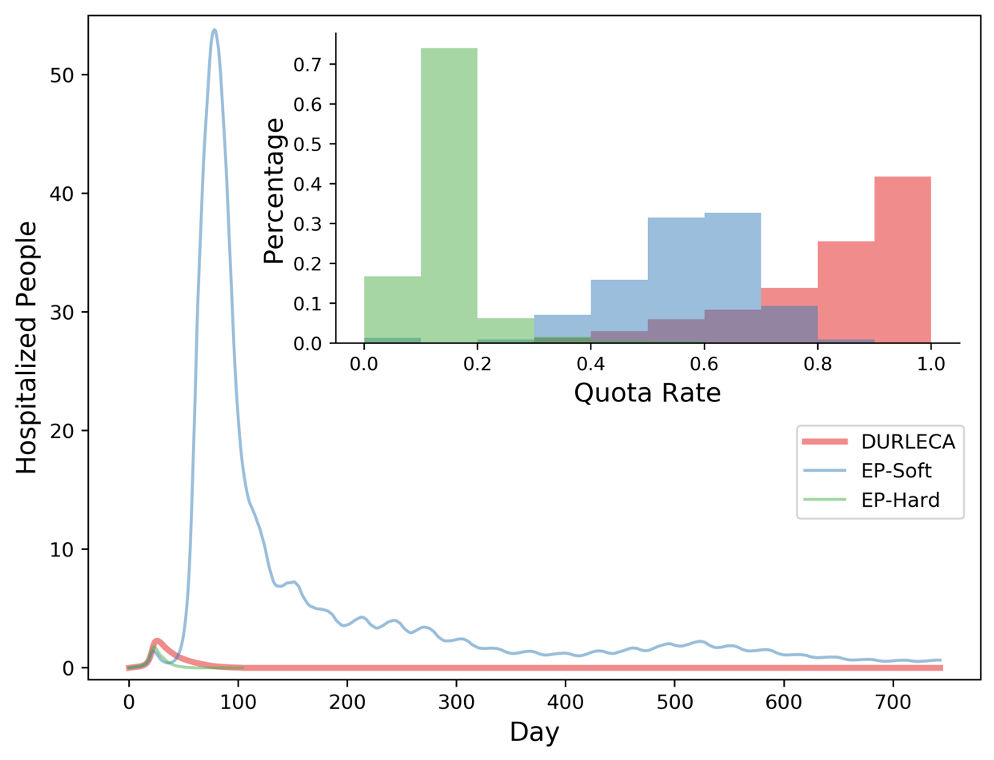}
    \caption{The simulation visualization of DURLECA and selected baselines when $t_{start}=20$. Main Figure: The number of hospitalized people along with time. Upper Right: The histogram for $Q_i^\tau$. Best viewed in color.}
    \label{fig:results}
\end{figure}

\begin{table}[ht]
\scalebox{0.95}{
    \centering 
        \begin{tabular}{l|lllll}
        \hline
        \hline 
          & Mean/Max $H$ & Total $R$ & $Q$ & $T_{20\%}^c$  & $T_{20\%}^r$  \\
        \hline
        
        No Intervention   & 27.21/157.55 & 1069.29 & 1 & 0 & 0 \\ \hline
        EP-Fixed $20\%$   & 4.03/17.96 & 877.32  & 0.20 & 724 & 724 \\
        EP-Fixed $15\%$    & 0.44/1.55 & 10.18  & 0.15 & 724 & 724 \\ 
        EP-Soft & 4.66/53.80 & 1040.68 & 0.57 & 9 & 19\\ 
        EP-Hard  & 0.45/1.78 & 8.31 & 0.13 & 36 & 36 \\ 
        EP-Lockdown  & 0.41/1.42  & 5.75   & 0 & 27 & 27 \\ \hline

        DURLECA  & 0.60/2.28 & 19.07 & 0.76 & 0 & 0 \\
        \hline
        \hline
        \end{tabular}
        \vspace{-0.7cm}
    }
    \caption{The simulation results of DURLECA and all baselines when $t_{start}=20$.}
\label{tab:results}
\vspace{-0.5cm}
\end{table}

\begin{figure}[t]
    \centering
    \includegraphics[width=0.53\textwidth]{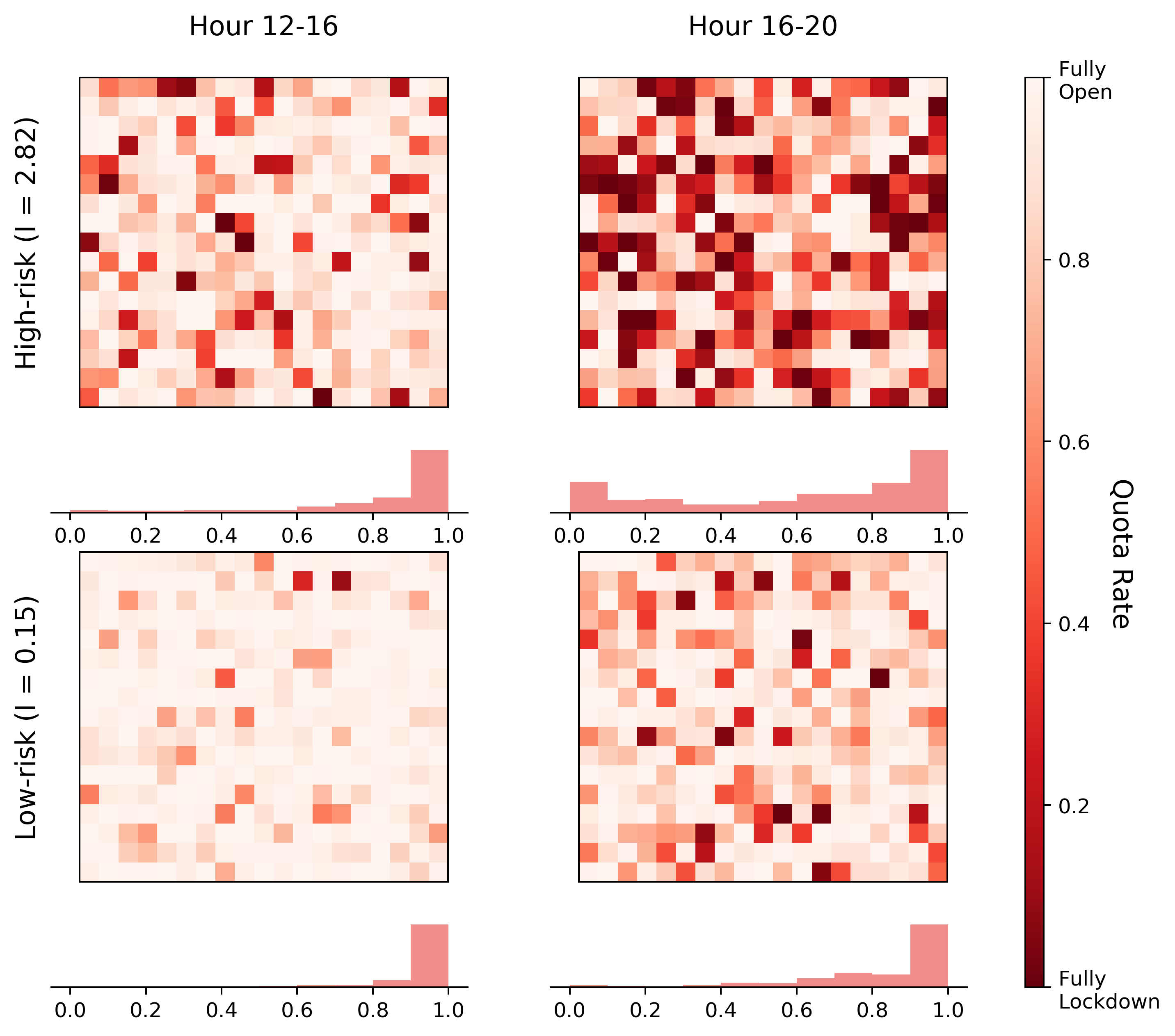}
    \caption{The spatial distribution and histogram of $Q_i^\tau$ given by DURLECA. We select four periods from a "high-risk day" and a "low-risk day". Each grid in one of the four 17$\times$19 maps represents a region in Beijing. Each histogram summarizes the distribution of quota rates in the respective period.} 
    \label{fig:quota}
    \vspace{-0.3cm}
\end{figure}

\textbf{Results and Analysis:}
We compare DURLECA with all baselines when $t_{start} = 20$ in Table~\ref{tab:results}. We also visualize three selected results in Figure~\ref{fig:results}. Expert baselines can achieve only one goal in the life-or-economy dilemma, while DURLECA can achieve both. 

EP-Soft can retain 57\% of the total mobility. However, it leads to an epidemic outbreak, reflected by the super-high value of Mean/Max $H$ and Total $R$. The healthcare system will break down. 
EP-Fixed ($X_q=15\%$), EP-Hard and EP-Lockdown can keep Mean/Max $H$ at a low level so that the healthcare system will not be overwhelmed. However, the low value of $Q$ indicates that all of them fail to retain mobility. The large value of $T_{20\%}^c$ and $T_{20\%}^r$  demonstrates that some regions and the city have to experience long-term lockdown, which is an unacceptable damage to the economy. Besides, the differences of EP-Fixed (15\%) and EP-Fixed (20\%) in Mean/Max $H$ and Total $R$ also indicate that the expert control is very vulnerable to mobility perturbation. The above results also manifest that all those expert policies fail at resolving the life-or-economy dilemma of epidemic control. 


Compared with those baselines, DURLECA simultaneously suppresses the epidemic and retains a large amount of mobility. DURLE-CA achieves low values of Mean/Max $H$, which guarantee the demands for hospitalization will not exceed the capacity of most countries. DURLECA also suppresses the total infected population at a low level, about $1\%$ of the total population. The red curve in Figure \ref{fig:results} presents the performance of DURLECA in epidemic suppression.

DURLECA also retains the most mobility. 76\% of the total mobility during the intervened period is retained. Furthermore, no regions will be fully locked down. DURLECA retains 70-100\% mobility for most regions in the city. The economic loss due to epidemic control can be significantly reduced. In all, DURLECA successfully resolves the life-or-economy dilemma.


DURLECA's control is highly customized and dynamic, which is hard to be mimicked by human experts. In Figure \ref{fig:quota}, we visualize the spatial distribution of quota rates and the associated histogram in four selected periods. Figure \ref{fig:quota} manifests that DURLECA's smartness in distributing quotas according to both epidemic risks and mobility patterns. The agent tends to give more quotas in a low-risk and low-mobility period and give fewer quotas in either a high-risk or a high-mobility period.   



\subsection{Comparison of the Scenarios of  Early/Late Intervention}

To examine whether DURLECA is still effective if the government's intervention is later than the discovery of the first patient, we compare DURLECA's performance in three scenarios. 
We have discussed the very-late-intervention scenario where the government starts to act 20 days after discovering the first patient ($t_{start}=20$) in Section 4.3. Here, we compare the early-intervention scenario ($t_{start}=0$) and the late-intervention scenario ($t_{start}=10$). The results are shown in Table \ref{tab:early_late}.


We find that: 
1) EP-Soft can control the epidemic in the early-intervention scenario. Because the virus has not widely spread, restricting the few infected areas is enough for epidemic suppressing. However, it fails to avoid an epidemic outbreak in the late-intervention scenario.
2) EP-Hard and EP-Lockdown can control the epidemic under both scenarios. However, it will lock all risky regions down and cut off most mobility. 
3) DURLECA successfully suppresses the epidemic while retains the majority of urban mobility in both scenarios.

\begin{table}[ht]
    \centering 
    \scalebox{0.95}{
        \begin{tabular}{l|l|lll}
        \hline
        \hline 
         & $t_{start}$ & Mean/Max $H$ & Total $R$ & $Q$ \\
        \hline
        EP-Soft    & 0  & 0.02/0.03 & 0.08 & 0.18 \\
        EP-Hard    & 0  & 0.03/0.04 & 0.10 & 0.13 \\
        EP-Lockdown    & 0  & 0.01/0.03 & 0.09 & 0 \\
        DURLECA    & 0  & 0.03/0.04 & 0.27 & 0.73 \\ \hline
         
        EP-Soft    & 10 &  4.67/62.10 & 1041.64 & 0.57 \\
        EP-Hard    & 10  & 0.08/0.14 & 0.64 & 0.11 \\
         EP-Lockdown    & 10 &  0.08/0.14 & 0.55 & 0 \\
         DURLECA    & 10 &  0.07/0.16 & 1.31 & 0.74 \\ 
        \hline
        \hline
        \end{tabular}
    }
    \caption{The simulation results of DURLECA and three baselines when $t_{start}=0,10$.}
\label{tab:early_late}
\vspace{-0.6cm}
\end{table}

\subsection{Generalization Ability}
We examine the generalization ability of DURLECA under different urban settings and diseases. Cities have different capacities for hospitalization treatment. Heterogeneous economic structures also cause cities' divergent tolerance for mobility restrictions. We vary the setting of $\{H_0, L_0\}$, which represents the change of urban features, and examine DURLECA's performance. The results are shown in Table~\ref{tab:h0l0}. The results demonstrate that DURLECA can find out different policies responding to the change of urban settings. For instance, we find that a higher $H_0$ leads to more mobility and more hospitalizations, which suggests that cities with higher hospitalization capacities can take more patients and retain more mobility. We also examine DURLECA's adaptiveness to various diseases. We vary the setting of $\{\beta_m$, $\beta_s$, $\gamma$, $\theta\}$ to simulate different diseases with different $R0$.
We find that DURELCA is also able to adjust epidemic-control policy to adapt to different diseases (Table \ref{tab:r0}). For instance, DURLECA provides loose mobility restrictions on low-$R0$ diseases but stringent mobility restrictions on high-$R0$ ones. DURLECA's adaptiveness to urban-setting and disease-setting changes not only demonstrates its generalization ability but also its smartness.

\begin{table}[ht]
    \centering 
    \scalebox{1}{
        \begin{tabular}{ll|llll}
        \hline
        \hline 
         $H_0$ & $L_0$ & Mean/Max $H$ & Total $R$ & $Q$ \\
        \hline
         1 & 72  & 0.54/1.74 & 10.02 & 0.38 \\ 
         3 & 72  & 0.60/2.28 & 19.07 & 0.76 \\
         10 & 72  & 2.79/6.34 & 223.29 & 0.90 \\ \hline
         3 & 48  & 1.69/4.60 & 153.54 & 0.88 \\ 
         3 & 72  & 0.60/2.28 & 19.07 & 0.76 \\
         3 & 168 & 0.45/1.58 & 16.69 & 0.71 \\
        \hline
        \hline
        \end{tabular}
    }
    \caption{The simulation results of DURLECA with different $H_0,L_0$ when $t_{start}=20$.}
\label{tab:h0l0}
\vspace{-0.4cm}
\end{table}

\begin{table}[ht]
    \centering 
    \scalebox{1}{
        \begin{tabular}{l|lll}
        \hline
        \hline 
         $R_0$ & Mean/Max $H$ & Total $R$ & $Q$ \\
        \hline
         1.4  & 0.35/1.20 & 10.96 & 0.86 \\ 
         2.1  & 0.60/2.28 & 19.07 & 0.76 \\ 
         3.5  & 0.81/4.08 & 23.18 & 0.46\\ 
        \hline
        \hline
        \end{tabular}
    }
    \caption{The simulation results of DURLECA under epidemics with different $R0$.} 
\label{tab:r0}
\vspace{-0.8cm}
\end{table}
\section{Related work}
\textbf{Epidemic Modeling:}
The SIR model is a widely used mathematical model in epidemiology, which divides the population into three states: susceptible, infected and recovered~\cite{SIR}. Based on the SIR model, Ogren and Martin used an embedded Newton algorithm to help find an optimal control strategy \cite{ogren2000optimal}. The distributed delay and discrete delay of SIR was also studied \cite{mccluskey2010complete}. Considering a more practical epidemic scenario, the SEIR model added an Exposed state to deal with the incubation period \cite{SEIR}. Others also strengthened the differential equations considering vaccination consequences for a measles epidemic \cite{allen10introduction}.
Later works also tried to incorporate human spatial patterns into the epidemic model. Sattenspiel et al. presented how contacts occur between individuals from different regions and how they influence epidemic spreads~\cite{sattenspiel1995structured}. 
Balcan et al. presented the GLobal Epidemic and Mobility model, which integrated sociodemographic and population mobility data in a spatially structured stochastic approach~\cite{balcan2010modeling}.
Different from previous works, we distinguish visible and invisible infections and model epidemic transmission upon traffic flows, so that to support exploring mobility-control policies for epidemic control.

\textbf{Graph Neural Network for OD-flows:}
The problem of estimating, predicting and controlling human flows between regions has been addressed using neural networks since \cite{lorenzo2013od}. Especially due to the reason that most OD flows are modeled based on graphs, 
Graph Neural Network (GNN) shows great importance and was first suggested in \cite{scarselli2009the}. Later GNNs were used to predict future mobility flows~\cite{chai2018bike, geng2019spatiotemporal, wang2019origin}. Besides, \cite{epidemic_gcn} borrowed knowledge from epidemic models to design GNN for node prediction in documents.
However, existing GNN architectures lack the ability to model the virus-spreading flow. Our designed Flow-GNN allows our model to characterize the virus-spreading flow and guarantees DURLECA's capability. 


\textbf{Deep Reinforcement Learning:}
Deep Reinforcement Learning (DRL) has been proved to be effective for control problems that have a large action space \cite{DQN, DDPG, double_q}. DQN \cite{DQN} and DDPG \cite{DDPG} are two representative DRL algorithms, proposed for discrete control problems and continuous control problems, respectively. 
To enable the agent to find an optimal solution, later works proposed to enhance exploration \cite{param_noise, explore_1, explore_2}. Imitation learning is another area of RL, where the goal is to enable the agent to behave like a human expert \cite{imitation}. AlphaGo proposed to start from imitation but further explore to outperform expert \cite{go}. In \cite{graph_protection}, DQN was also used for node protection against epidemic under a single objective. Compared with it, both our control action and objectives are more complex and practical, and thus our RL training are more challenging. We design two strategies to address the exploration challenge.

\section{Future research and Conclusion}


A series of problems ask for future study on smart-and-privacy-protected epidemic control while DURLECA can be the framework. In this research, we do not consider the uncertainty of mobility and epidemic information when DURLECA explores epidemic policies. It asks for future work to explore the algorithm for searching a robust policy when the information is uncertain. A practical policy has to be robust even if there exist errors in the input data. 

To conclude, this research demonstrates a sequence of important facts, which broaden the vision of human society for epidemic control and are listed below:
\begin{itemize}
    \item Private data is dispensable because restricting the aggregated inter-regional mobility sufficiently lowers the probability of infectious people's movement and thus suppresses the risk of epidemic transmission.
    \item Resolving the life-or-economy dilemma of epidemic control must allow dynamic and customized regional policies.
    \item The powerfulness of our GNN-enhanced RL in epidemic control manifests that field knowledge is critical for AI-system architecture and valuable for neural network training.  
\end{itemize}
In all, smart governance empowered by AI will protect future society from the loss of lives due to epidemics and the economic risk caused by epidemic control. 






\section*{Acknowledgement}
This work was supported in part by The National Key Research and Development Program of China under grant 2018YFB1800804, the National Nature Science Foundation of China under U1936217,  61971267, 61972223, 61941117, 61861136003, Beijing Natural Science Foundation under L182038, Beijing National Research Center for Information Science and Technology under 20031887521,  research fund of Tsinghua University - Tencent Joint Laboratory for Internet Innovation Technology, and The Information Core Technology Center at Institute for Interdisciplinary.

\bibliographystyle{ACM-Reference-Format}
\bibliography{cite}

\newpage
\appendix
\section{Notation Summary}

\begin{table}[ht]
\centering
\scalebox{1}{
\begin{tabular}{l|l}
\hline
\hline 
Term/Notation & Definition \\
\hline
superscript $\tau$  & At time step $\tau$. \\ 
subscript $d$  & The original demand without restrictions. \\
subscript $p$  & With restriction $p$. \\
subscript $i$, $j$ & Region index \\ \hline

$M_*^\tau$  & The mobility. A matrix.\\ 
$M_{*,i,j}^\tau$  & The OD flow from $i$ to $j$. A scalar.  \\
$M_{*,i}^\tau$  & $\sum_j M_{*,i,j}^\tau$. The out-flow from $i$. A scalar.  \\ 
$\overline{M_{*,i}}$ &  $\frac{1}{T}\sum_\tau M_{*,i}^\tau$. The mean out-flow from $i$. A scalar. \\ \hline
$Q_i^\tau$ & $\frac{M_{p,i}^\tau}{M_{d,i}^\tau}$. The region quota rate. A scalar. \\
$Q^\tau$ & $\frac{\sum_i M_{p,i}^\tau}{\sum_i M_{d,i}^\tau}$. The city quota rate. A scalar. \\
$Q$ & $\frac{\sum_\tau \sum_i M_{p,i}^\tau}{\sum_\tau \sum_i M_{d,i}^\tau}$. The total quota rate. A scalar. \\
\hline
\hline
\end{tabular}
}
\caption{The summary of mobility-related notations. Subscript $*$ can be either $d$ or $p$.}
\label{tab:definitions}
\end{table}

\section{Experiment Settings and Reproducibility}

\subsection{Dataset} 
We list the dataset details in Table \ref{tab:dataset}, where $P_{move}$ counts the mean probability for an individual to move in one hour. 
\begin{table}[ht]
\centering
\begin{tabular}{lllll}
\hline
\hline 
City & Regions & Mean Population & $P_{move}$ & Duration\\
\hline
Beijing & $17\times19$ & 1686 & 0.18 & 744 Days\\
\hline
\hline
\end{tabular}
\caption{The summary of the prolonged dataset.}
\label{tab:dataset}
\end{table}

\textbf{Privacy and ethical concerns:} 
We have taken the following procedures to address privacy and ethical concerns. First, all of the researchers have been authorized by the data provider to utilize the data for research purposes only. Second, the data is completely anonymized. Third, we store all the data in a secured off-line server. 

\subsection{Implementation Details}
Without the loss of generality, we set the moving transmission rate $\beta^m = \frac{3}{24}$, the staying transmission rate $\beta^s = \frac{0.1}{24}$, the hospitalized rate $\gamma=\frac{0.3}{24}$ and the recover rate $\theta=\frac{0.3}{24}$. Without intervention, the estimated basic reproduction rate $R_0$ is 2.1 at the initial stage of the epidemic. 
For the reward, we mainly set $\lambda = 0.99, L_0 = 72, H_0 = 3, k_h = 1$. For the pseudo expert, we set $X_{h} = 1, X_{l} = 168$. For the extreme point threshold, we set $I_{t} = 100, L_{t} = 336$.

During training, we randomly initialize an epidemic state at the start of each episode. We train DURLECA for 400,000 steps, using Adam optimizer with the learning rate as 0.0001.
During testing, we fix one epidemic-initialization setting and compare different baselines. Considering the randomness of training, we train DURLECA with different random seeds 5 times for each set of configurations, and choose the one that achieves the best episode reward to report as the result.

We mainly implement DURLECA based on Keras-RL \cite{keras_rl} with our modifications.

\subsection{Disease $R0$} 
In a classical SIR model, the basic reproduction rate $R0$ is calculated as $R0=\frac{\beta}{\gamma}$ \cite{R0}. In our model, as the infection has been divided into two parts, we estimate an averaged $\overline{\beta}$ over $\beta_m$ and $\beta_s$ according to their corresponding population size, 
\begin{align}
    \overline{\beta} = P_{move}\beta_m + (1-P_{move})\beta_s.
\end{align}
Then we estimate $R0=\frac{\overline{\beta}}{\gamma}$.

In Section 4.5, we vary the setting of $\{\beta_m, \beta_s\}$ and keep $\{\gamma, \theta\}$ the same. To make a fair comparison, we also vary $t_{start}$ for each simulated disease to make sure the city has nearly the same number of hospitalized people when we start the intervention. For $R0=1.4$, we set $\{\beta_m  = 1.9, \beta_s = 0.1, t_{start} = 45\}$. For $R0=2.1$, we set $\{\beta_m  = 3, \beta_s = 0.1, t_{start} = 20\}$. For $R0=3.5$, we set $\{\beta_m  = 5, \beta_s = 0.2, t_{start} = 10\}$.

\section{Ablation Study}

To evaluate the effectiveness of our proposed Flow-GNN and RL exploration strategies, we conduct ablation studies in this section. 

\textbf{GNN-Baselines}: To evaluate the effectiveness of our proposed Flow-GNN, we use the well known GraphSageConv layer \cite{graphsage} and our modified GraphSageConv layer to replace the proposed Flow-GNN layer in the actor network and the critic network. We name the two baselines as \textbf{GNN-Mean} and \textbf{GNN-Softmax}. 

The layer calculation of GNN-Mean follows Equation~(\ref{eq:GNN-mean}):
\begin{align}
    & f_i^k = \sigma (W^k (f_{i}^{k-1}, \frac{1}{N(i)}\sum_{j\in N(i)} f_{j}^{k-1}) + B^k), \label{eq:GNN-mean}
\end{align}
where $N(i)$ denotes the connected regions of $i$.

The layer calculation of GNN-Softmax follows Equation~(\ref{eq:GNN-softmax}):
\begin{align}
    & w_{j,i} = \frac{exp(M_{*,j,i}^{\tau+k-1})}{\sum_{j\in N(i)}exp(M_{*,j,i}^{\tau+k-1})}, \notag \\
    & f_i^k = \sigma (W^k (f_{i}^{k-1}, \sum_{j\in N(i)} w_{j,i} f_{j}^{k-1}) + B^k). \label{eq:GNN-softmax} 
\end{align}
    
\textbf{RL-Baselines}: To evaluate the effectiveness of our RL exploration strategies, we remove the pseudo-expert strategy and the avoiding-extreme-points strategy, respectively. We refer to the two baselines as \textbf{RL-NoEP} and \textbf{RL-NoThre}.

\begin{table}[ht]
    \centering 
    \scalebox{0.9}{
        \begin{tabular}{l|lllll}
        \hline
        \hline 
          & Mean/Max $H$ & Total $R$ & $Q$ & $T_{20\%}^c$  & $T_{20\%}^r$ \\
        \hline
        
        No Intervention  & 27.21/157.55 & 1069.29 & 1 & 0 & 0 \\ \hline
    
        GNN-Mean & -/- & - & - & - & - \\
        GNN-Softmax  & 0.53/1.88 & 7.79 & 0.06 & 26 & 28 \\
        
        RL-NoEP  & 0.41/1.45 & 5.87 & 0.00 & 27 & 27 \\
        RL-NoThre  & 1.43/3.68 & 86.82 & 0.75 & 0 & 9 \\ \hline
        
        DURLECA & 0.60/2.28 & 19.07 & 0.76 & 0 & 0 \\
        \hline
        \hline
        \end{tabular}
    }
    \caption{Ablation study when $t_{start}=20$.}
\label{tab:ablation}
\end{table}

\textbf{Results and Analysis:}
As shown in Table \ref{tab:ablation}, without Flow-GNN or the proposed RL exploration strategies, the agent fails to learn a good policy. 

The failure of GNN-Mean comes from its inability to learn weighted edge information, i.e., how many people move from one region to another. With considering weighted edge information, GNN-Softmax still fails to retain mobility, as it can not describe traffic flows and the epidemic transmission upon it. These prove the effectiveness of our proposed Flow-GNN.

RL-NoEP gives a long-term lockdown to the whole city, which is a typical local optimum. As for RL-NoThre, the agent successfully finds one policy that achieves relatively low hospitalizations and high mobility. However, this solution is worse than DURLECA. Besides, we find that the success of RL-NoThre highly relies on luck. During our five repeating experiments, the agent was stuck in local optimums for four times, giving a long-term lockdown to the whole city. These, as discussed earlier in Section 1, are due to the difficulty of exploration. The agent is easy to encounter extreme points during exploration, and the extreme points force the agent to adopt conservative policies, i.e., lock the whole city down. Compared with RL-NoEP and RL-NoThre, DURLECA is guided by a pseudo-expert and is designed to avoid extreme points. Thus, DURLECA can find much better solutions.

\end{document}